\documentclass[letterpaper]{article} 
\usepackage{aaai25}  
\usepackage{times}  
\usepackage{helvet}  
\usepackage{courier}  
\usepackage[hyphens]{url}  
\usepackage{graphicx} 
\usepackage{natbib}  
\usepackage{caption} 
\usepackage{algorithm}
\usepackage{algorithmic}
\usepackage{booktabs}
\usepackage{amsmath} 
\usepackage{amssymb}
\usepackage{amsfonts}
\usepackage{times}  
\usepackage{helvet}  
\usepackage{courier}  
\usepackage[hyphens]{url}  
\usepackage{graphicx} 
\usepackage{natbib}  
\usepackage{caption} 
\usepackage{textcomp}
\usepackage{graphicx}
\usepackage{amsmath}
\usepackage{amsfonts} 
\usepackage{multirow, graphicx}
\usepackage{colortbl}
\usepackage{bbding}
\usepackage{tabularx}
\usepackage{graphicx}
\usepackage{amsmath}
\usepackage{booktabs}
\usepackage{dsfont}
\usepackage{amsfonts,amssymb}
\usepackage{graphics}
\usepackage{tikz}
\usepackage{ragged2e}

\usepackage{algorithm}
\usepackage{algorithmic}

\usepackage{newfloat}
\usepackage{listings}
\usepackage{multirow}
\usepackage{amssymb}  
\usepackage{array}      
\usepackage{newfloat}
\usepackage{listings}
\DeclareCaptionStyle{ruled}{labelfont=normalfont,labelsep=colon,strut=off} 
\floatstyle{ruled}
\newfloat{listing}{tb}{lst}{}
\floatname{listing}{Listing}
\title{Neighbor Does Matter: Density-Aware Contrastive Learning \\ for Medical Semi-supervised Segmentation}
\author{
    Feilong Tang\textsuperscript{\rm 1}\equalcontrib,
    Zhongxing Xu\textsuperscript{\rm 1}\equalcontrib,
    Ming Hu\textsuperscript{\rm 1},
    Wenxue Li\textsuperscript{\rm 1},
    Peng Xia\textsuperscript{\rm 3}, \\
    Yiheng Zhong\textsuperscript{\rm 2}, 
    Hanjun Wu\textsuperscript{\rm 2},
    Jionglong Su\textsuperscript{\rm 2}\textsuperscript{\dag},
    Zongyuan Ge\textsuperscript{\rm 1}\thanks{Corresponding author: Jionglong Su; Zongyuan Ge}
}

\affiliations{
    \textsuperscript{\rm 1}AIM Lab, Faculty of IT, Monash University\\
    \textsuperscript{\rm 2}Xi’an Jiaotong-Liverpool University\\
    \textsuperscript{\rm 3}UNC-Chapel Hill \\
    Feilong.Tang@monash.edu
}

\usepackage{bibentry}

\begin{document}

\maketitle

\begin{abstract}
In medical image analysis, multi-organ semi-supervised segmentation faces challenges such as insufficient labels and low contrast in soft tissues. To address these issues, existing studies typically employ semi-supervised segmentation techniques using pseudo-labeling and consistency regularization. However, these methods mainly rely on individual data samples for training, ignoring the rich neighborhood information present in the feature space. In this work, we argue that supervisory information can be directly extracted from the geometry of the feature space. Inspired by the density-based clustering hypothesis, we propose using feature density to locate sparse regions within feature clusters. Our goal is to increase intra-class compactness by addressing sparsity issues. To achieve this, we propose a \textit{Density-Aware Contrastive Learning (DACL)} strategy, pushing anchored features in sparse regions towards cluster centers approximated by high-density positive samples, resulting in more compact clusters. Specifically, our method constructs density-aware neighbor graphs using labeled and unlabeled data samples to estimate feature density and locate sparse regions. We also combine label-guided co-training with density-guided geometric regularization to form complementary supervision for unlabeled data. Experiments on the Multi-Organ Segmentation Challenge dataset demonstrate that our proposed method outperforms state-of-the-art methods, highlighting its efficacy in medical image segmentation tasks.
\end{abstract}

%

\section{Introduction}
Multi-organ segmentation in medical images is essential for many clinical applications, such as computer-aided intervention~\cite{qu2024abdomenatlas,tang2023duat,wang2022stepwise,xu2024polyp,ye2022SSL,ye2024SSL,chen2023DG,chen2023UDA,chen2024CTTA}, surgical navigation~\cite{zheng2024federated,hu2024ophclip,hu2025ophnet,tang2024discriminating}, and radiotherapy~\cite{mofid2024deep}. However, complex morphological structures of multiple organs, low contrast of soft tissues, and blurred boundaries between adjacent organs make dense annotation of multi-organ images extremely expensive and challenging. The scarcity of annotations has become a bottleneck that limits the performance of deep learning-based multi-organ segmentation models. As a solution, semi-supervised learning (SSL)~\cite{huang2024combinatorial,chi2024adaptive,liu2024rolling,ye2024CSSL} enhances labeling efficiency by leveraging a limited amount of labeled data alongside a large amount of unlabeled data. SSL aims to extract effective training signals from unlabeled images to enhance model generalization.

\begin{figure*}[t]
  \centering
  \begin{tabular}{cc}
    \includegraphics[width=0.95\textwidth]{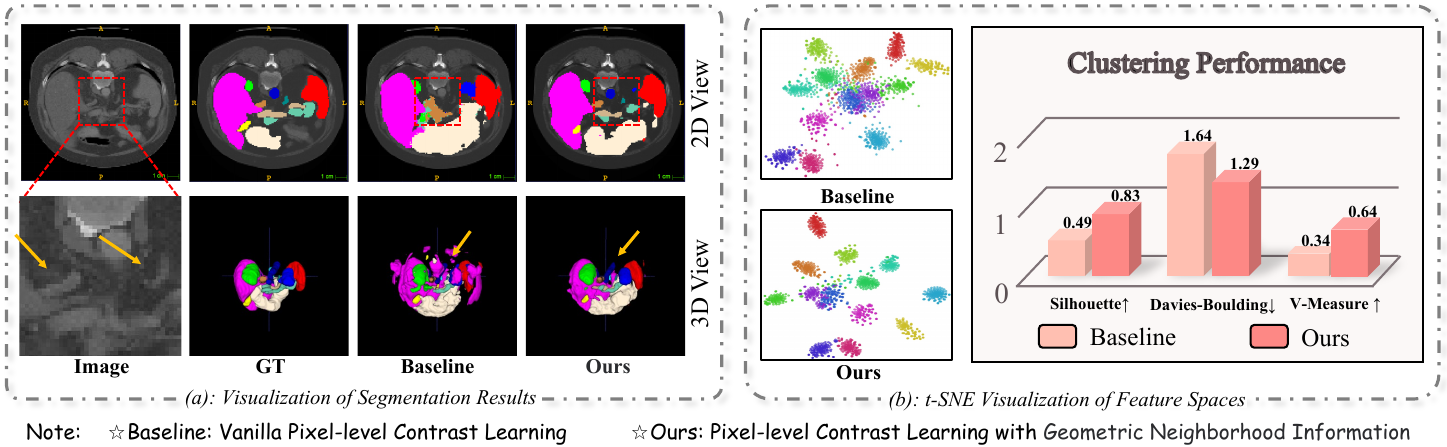} 
  \end{tabular}
  \caption{Left (a): Qualitative comparison of abdominal organs. As observed in the red boxes, the contrastive learning method that incorporates geometric neighborhood information achieves a more plausible appearance and a more complete structure compared to the conventional SSL method. Right (b): t-SNE~\cite{van2008visualizing} reveals that our method produces more compact class clusters with clearer separation between each class compared to the baseline. The clustering performance is evaluated using Silhouette~\cite{rousseeuw1987silhouettes}, Davies-Boulding~\cite{davies1979cluster} and V-Measure~\cite{rosenberg2007v} score. Our method outperforms the baseline in clustering performance metrics.}
  \label{fig1}
\end{figure*}

To handle unlabeled data, most existing SSL methods are based on the common density clustering assumption~\cite{wang2023hunting}, \textit{i.e.,} samples located in the same high-density area are likely to belong to the same class, and the decision boundary should be located in the low-density area. In other words, by achieving inter-class separation based on density, semi-supervised models can output highly confident predictions without the supervision of corresponding labels. Current multi-organ segmentation methods~\cite{wong2023scribbleprompt} mainly focus on separating low-density areas. These methods make invariant predictions to perturbations around each unlabeled data point or encourage nearby samples to make the same predictions, including performing consistency regularization~\cite{chen2023magicnet} or generating pseudo-labels~\cite{basak2023pseudo}. Despite making some progress, current methods mainly utilize single data samples and ignore the rich neighborhood information for learning more discriminative features. Therefore, these methods face challenges with regards to data scarcity and feature sparsity.


Recent advances in manifold learning have established two properties~\cite{dosovitskiy2020image}: (1) allowing similar features to be closer to each other, while dissimilar features move further apart; (2) providing superior manifold metrics to measure discrepancies with different statistical and geometrical properties. Based on these, we mine effective neighborhood information from the geometry of the feature space. As shown in Fig.~\ref{fig1} (a), learning that incorporates neighborhood information can effectively capture the complex shapes of various organs (\textit{e.g.,} large liver and small pancreas) and produce more compact and discriminative embeddings (Fig.~\ref{fig1} (b)). 
In contrast, directly using pixel-level features in contrastive learning methods can misidentify some organ pixels as background, particularly for smaller organs (Fig.~\ref{fig1} (a)).

In this work, we propose a learning strategy called \textbf{D}ensity-\textbf{A}ware \textbf{C}ontrastive \textbf{L}earning (DACL). Within the feature clusters of both labeled and unlabeled data, we construct density-aware neighbor graphs by measuring local density based on the average distance between the target feature and its nearest neighbors. To achieve robust perception, we introduce a category memory bank to overcome the limitations of small batches, enabling global estimation of intra-class density in a feature-to-bank manner. Simultaneously, we use density-aware maps to comprehensively estimate the final density and locate sparse regions. We also propose \textbf{S}oft \textbf{D}ensity-Guided \textbf{C}ontrastive \textbf{L}earning (SDCL) to explicitly shrink sparse regions for more compact clusters. Specifically, sparsity search is conducted within class feature clusters using density-aware neighbor graphs to locate low-density features as anchors. Meanwhile, features from dense areas are selected as positive keys to approximate cluster centers. Subsequently, we calculate the positiveness score of each low-density feature with the corresponding positive key and incorporate it into the contrastive loss calculation. A lower score indicates greater sample distance. Through contrastive learning, anchors are drawn closer to positive keys, enhancing feature clustering.

We evaluate our proposed method on Automatic Cardiac Diagnosis Challenge (ACDC)~\cite{bernard2018deep} and the Synapse multi-organ segmentation~\cite{landman20152015} under various semi-supervised settings, where our method achieves state-of-the-art performances. The main contributions of our work are summarized as follows:


\begin{itemize}
\item We propose a Density-Aware Contrastive Learning (DACL) strategy that focuses on effectively utilizing the geometric structure of the feature space to address the challenge of data scarcity and feature sparsity in semi-supervised multi-organ segmentation. 

\item We propose mining effective neighborhood information in the feature space and integrating it into contrastive learning, consequently enhancing intra-class compactness in the feature space. 

\item Experimental results on benchmark datasets demonstrate that our method significant improves upon the efficacy of previous state-of-the-art methods.
\end{itemize}

\section{Related Work}
\textbf{Semi-supervised Medical Image Segmentation.} Recent efforts in semi-supervised segmentation have been focused on incorporating unlabeled data into CNNs~\cite{tang2024hunting,zhao2024sfc,liu2022copy,wu2024pose,wu2024joint,yang2023action}, which can be broadly categorized into five groups: self-training~\cite{chi2024adaptive}, co-training~\cite{ma2024constructing}, deep adversarial learning~\cite{xu2024deep} and self-ensembling ($\pi$-model~\cite{deng2024semi} and Mean-Teacher model~\cite{zhao2023alternate}). For instance, Liu et al.~\cite{liu2023temporal} achieved Deep Co-Training by learning two classifiers on two views. Liu et al.~\cite{liu2024diffrect} designed a Deep Adversarial Network, which enforced the segmentation of unlabeled data to be similar to the labeled ones. Yu et al.~\cite{yu2019uncertainty} proposed an Uncertainty-Aware Mean-Teacher, which enables the student to learn from reliable targets. Nevertheless, considering most regions with low uncertainty tend to be easy for a segmentation network to make correct predictions~\cite{jiang2024ph123}, low prediction quality for hard regions such as boundaries remains a challenge for performance improvement~\cite{wu2022mutual}. Therefore, we aim to improve
the general quality of the learning target directly based on a co-training framework.

\begin{figure*}[t]
  \centering
  \begin{tabular}{cc}
    \includegraphics[width=0.95\textwidth]{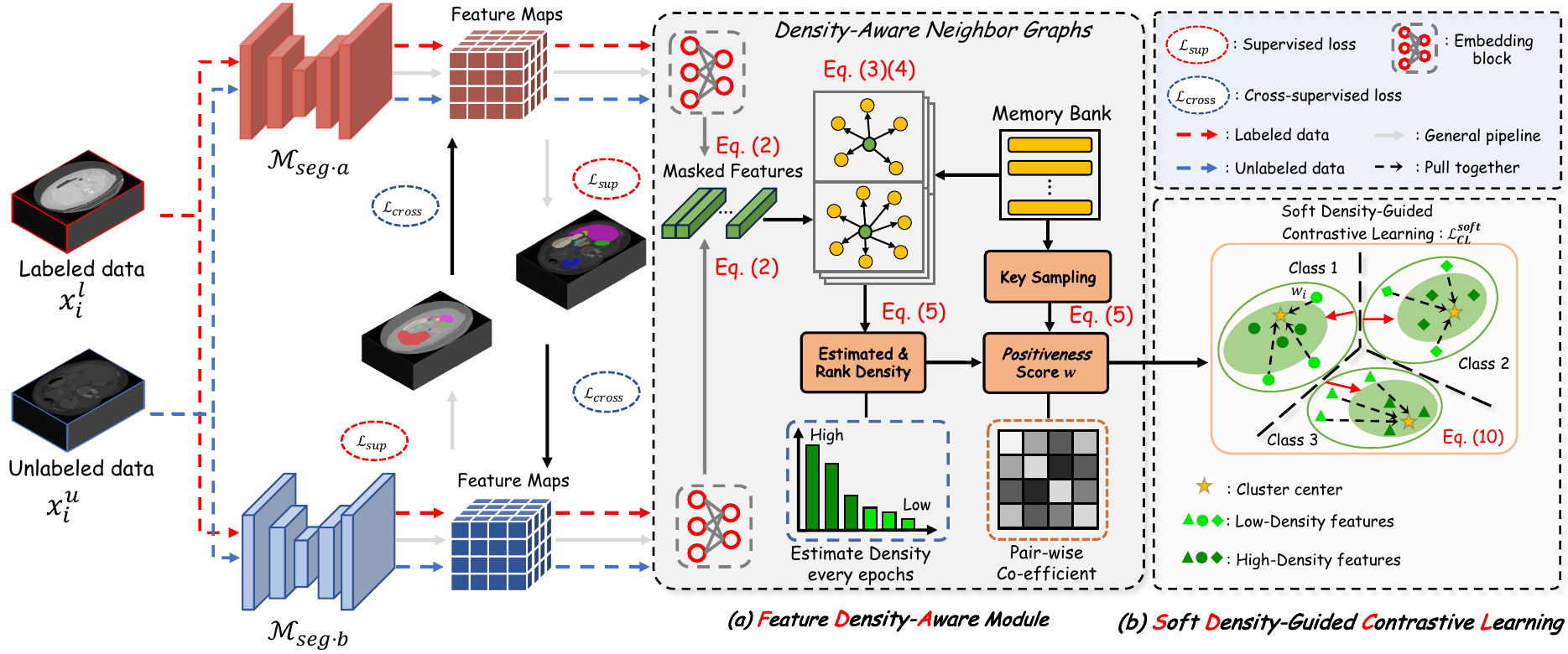} 
  \end{tabular}
  \caption{Overview of the proposed unified learning framework. (a) shows the feature density-aware module. (b) shows our Soft Density-guided Contrastive Learning strategy. For labeled image $x_i^l$, we apply the commonly-used supervised loss $\mathcal{L}_{sup}$ to update model parameters. For unlabeled image $x_i^u$, the model is optimized with the cross-supervised consistency loss $\mathcal{L}_{cross}$ and one manifold constraint $\mathcal{L}^{soft}_{CL}$ to explore the geometry of the feature space.}
  \label{method}
\end{figure*}

\noindent\textbf{Contrastive Learning.} Image-level contrastive learning has achieved significant results in self-supervised learning. Typical methods such as SimCLR~\cite{chen2020simple} have proposed pipelines for learning augmentation-invariant representations, which outperform traditional supervised pre-training methods~\cite{wu2024voco}. Building on this, we guide pixel-level feature learning by incorporating label information and transforming feature density into training signals, enhancing model performance.



\section{Methodology}
Fig.~\ref{method} illustrates the overall architecture of our proposed DACL framework, which builds upon the co-teaching paradigm. This framework aims to mine effective supervision from unlabeled images by leveraging a limited set of annotations. It employs a density-guided contrastive learning strategy to regularize the structure of clusters in the feature space. Within each mini-batch, we evaluate the density to locate features in sparse regions as anchors, while selecting features from dense regions as positive keys approximating the cluster center. By pushing low-density anchors towards the approximated cluster center, we aim to minimize contrastive loss, dynamically shrink cluster volume, and increase intra-cluster compactness. Furthermore, we softly calculate the positiveness score for each low-density feature with its corresponding positive key and incorporate it into the contrastive loss calculation.

\subsection{Co-training Framework with Geometric Loss}

In a general semi-supervised learning problem, assume that the whole dataset consists of $N_L$ labeled samples $\left\{\left(x_i^l, y_i\right)\right\}_{i=1}^{N_L}$ and $N_U$ unlabeled samples $\left\{x_i^u\right\}_{i=1}^{N_U}$, where $x_i \in \mathbb{R}^{H \times W}$ denotes an image and $y_i \in \mathbb{R}^{N \times H \times W}$ is the ground-truth annotation with $N$ classes (including background). $W$ and $H$ respectively represent the width and height of the input image. We adopt a co-teaching paradigm such as MCNet~\cite{wu2021semi} to enforce two segmentation models $\mathcal{M}_{seg \cdot a}$ and $\mathcal{M}_{seg \cdot b}$ to predict the prediction results of its counterpart. To leverage limited annotations and extract effective supervisory information from unlabeled images, we identify sparse region features as anchors and use representative features as positive keys. Our objective is to minimize the contrastive loss by pushing low-density anchors towards the cluster center, enhancing cluster compactness. A unified loss function $\mathcal{L}$ optimizes the model:
\begin{equation}\small
\mathcal{L}=\mathcal{L}_{sup}+\lambda_{cross} \mathcal{L}_{cross}+ \lambda_{CL}\mathcal{L}^{soft}_{CL}, \label{eq1}
\end{equation}
where $\mathcal{L}^{soft}_{CL}$ is the soft density-guided contrastive loss, and $\mathcal{L}_{sup}$ and $\mathcal{L}_{cross}$ are the supervised loss and cross-supervised loss, respectively. $\lambda_{cross}$ and $\lambda_{CL}$ are used to balance the relationship between losses (specifically, we choose $\lambda_{cross}=1$ and $\lambda_{CL}$ as Gaussian warming up function $\lambda(t)=0.1 \times e^{-5\left(1-t / t_{\max }\right)^2}$, where $t$ is the current training step and $t_{\max }$ is the maximum training step). During training, a batch of mixed labeled and unlabeled samples are fed into the network. The supervised loss $\mathcal{L}$ is applied only to labeled data, while all samples are utilized to construct cross-supervised learning. Refer to the Appendix for details on $\mathcal{L}_{sup}$ and $\mathcal{L}_{cross}$ losses.

\subsection{Feature Density-Aware Module}
Before analyzing the density of features within each class, we first approximate the distribution of class features. The features of classes within images or batches can only offer a partial perspective. For each image $x$, the input to encoder extract feature maps $f\in \mathbb{R}^{W \times H \times D}$ with $D$ channels and the feature maps are mapped to the projection space $j=z(f)$ by a projection head $z$ for instance prototyping. Each instance embedding represents the regional semantics of the categories observed in $x$ based on the corresponding ground truth or pseudo labels $M_n$. Specifically, for the $n$-th category, its projected features is summarized to a vector $v_n \in \mathbb{R}^D$ by masked average pooling (MAP) \cite{siam2019amp}:
\begin{equation}\small
v_n =\frac{\sum_{x=1}^{W} \sum_{y=1}^{H} \textbf{T}_n(x,y) * j(x,y)}{\sum_{x=1}^{W} \sum_{y=1}^{H} \textbf{T}(x,y)}, 
\label{tau}
\end{equation}
where $\textbf{T}_n= \mathbf{1}\left({M_n}>\phi\right) \in \{0,1\}^{W \times H}$ is a binary mask, emphasizing only strongly-activated pixels of class $n$ in its activation map. $\mathbf{1}(\cdot)$ is an indicator function, and $\phi$ is a hyper-parameter denoting the threshold of the reliability score. $*$ represents element-wise multiplication. Here, $v_n$ is compact and lightweight, allowing feasible exploration of its relationships with numerous other samples.

We propose a dynamic memory bank to aggregate class features from the entire dataset, enabling more comprehensive modeling of class features. Specifically, we extract embeddings using two segmentation models. The embeddings $v \in \mathbb{R}^{B \times D}$ are projection features with batch size $B$. These embeddings are stored in the memory bank. For each class $n$, the feature memory $\mathcal{G}_n$ is updated following a First-In, first-Out strategy while maintaining a fixed size $L$. In this module, we adopt a feature-to-bank approach to construct density-aware graphs. Within a mini-batch, the features for class $n$ are denoted as $\mathcal{V}_n=\{v_i\}^{B}_{i=1}$. As shown in Fig.~\ref{method} (a), we measure the density of each $v$ by constructing a $k$-nearest neighbor graph $\mathrm{NN}(v)= \{v_j\}^{k}_{j=1}$ within $\mathcal{G}_n \cup \mathcal{V}$. The density $d_v$ is computed as the average cosine similarity between $v$ and its $k$ neighbors:
\begin{equation}\small
d_v=\frac{1}{|\mathrm{NN}(v)|} \sum_{r \in \mathrm{NN}(v)} \tilde{\mathbf{f}}_r^T \tilde{\mathbf{f}}_v,
\label{2}
\end{equation}
where $\tilde{\mathbf{f}}_r$, $\tilde{\mathbf{f}}_v$ are the L2-normalized feature embedding of $r$ and $v$. We compute densities $\{d_v\}$ to identify the low-density and high-density features of $v$. Meanwhile, batch features with density values $\{(v, d_v)\}$ are used to update the corresponding memory bank.
Since density estimation with a few neighbors tends to focus on local regions, true cluster centers fail to be captured. To achieve a more accurate estimation, we introduce multi-scale nearest neighbor graphs:
\begin{equation}\small
\mathrm{NN}_{\text{multi-scale}} = \{\mathrm{NN}_{k_i}\}^N_{i=1}
\end{equation}
for the target feature vector $v$, where $k_1<k_2<\ldots<k_n$ and $n$ denotes the number of different neighbor scales used. We compute density values $d^1, d^2, \ldots, d^n$ using different graphs according to Eq.~\ref{2}. The final estimation is calculated as the average density value: 
\begin{equation}\small
d_v=\frac{1}{N} \sum_{i=1}^N d_{v}^i. 
\end{equation}

\subsection{Soft Density-Guided Contrastive Learning}
Here, we introduce Soft Density-Guided Contrastive Learning for complementary supervision on unlabeled data. The feature vectors in each mini-batch are ranked based on their density values, and those with low-density features are selected as anchors. Meanwhile, features are extracted from a memory bank as positive keys, and the positiveness scores between low-density features and positive keys are computed and incorporated into the contrastive loss.

\begin{table*}[t]
\centering
\tiny
\resizebox{\textwidth}{!}{
\begin{tabular}{l|cc|c|c|c|cc|c|c|c}
\toprule[1pt]
\rowcolor{cyan!30}
&  \multicolumn{5}{c|}{\textbf{ACDC database}} &  \multicolumn{5}{c}{\textbf{Synapse dataset}} \\ \midrule
\multirow{4}{*}{{\textbf{Methods}}} & \multicolumn{2}{c|}{\textbf{Scans used}} & \multicolumn{3}{c|}{ \textbf{Metrics} } & \multicolumn{2}{c|}{\textbf{Scans used}} & \multicolumn{3}{c}{ \textbf{Metrics} } \\
\cmidrule(r){2-11} & Labeled & Unlabeled & Dice(\%)$\uparrow$ & JC(\%)$\uparrow$ & ASD(voxel)$\downarrow$ & Labeled & Unlabeled & Dice(\%)$\uparrow$ & JC(\%)$\uparrow$ & ASD(voxel)$\downarrow$ \\ \midrule 

V-Net & 3(5\%) & 0 & 43.30$\pm$8.6 & 37.01$\pm$6.90 & 12.41$\pm$0.8 & 2(10\%) & 0 & 12.37$\pm$9.7 & 10.29$\pm$6.8 & 68.14$\pm$6.9 \\

V-Net & 7(10\%)  & 0 & 79.62$\pm$1.5 & 68.11$\pm$1.88 & 2.94$\pm$1.2 & 4(20\%) & 0 & 32.13$\pm$6.4 & 27.58$\pm$3.4 & 35.84$\pm$2.2 \\

V-Net & 70(All\%) & 0 & 91.54$\pm$0.7 & 84.87$\pm$0.36 & 0.90$\pm$0.6 & 20(All\%) & 0 & 62.09$\pm$1.2 & 58.43$\pm$2.6 & 10.28$\pm$3.9 \\ \midrule

$\textnormal{CPS}$ & \multirow{10}{*}{{\textbf{3(5\%)}}} & \multirow{10}{*}{{\textbf{67(95\%)}}} & 57.67$\pm$0.5 & 44.02$\pm$0.6 & 3.75$\pm$1.8 & \multirow{10}{*}{{\textbf{2(10\%)}}} & \multirow{10}{*}{{\textbf{18(90\%)}}} & 18.07$\pm$1.2 & 14.42$\pm$0.9 & 57.64$\pm$1.8 \\

$\textnormal{MC-Net}$ &  &  & 61.56$\pm$1.9 & 49.45$\pm$0.8 & 2.38$\pm$0.4 &  & & 21.96$\pm$6.1 & 19.42$\pm$2.6 & 55.42$\pm$4.6 \\

$\textnormal{URPC}$ & & & 54.69$\pm$1.2 & 44.78$\pm$0.9 & 3.47$\pm$0.3 & & & 26.37$\pm$1.5 & 23.62$\pm$1.2 & 53.95$\pm$9.3 \\

$\textnormal{SS-Net}$ & & & 64.12$\pm$1.7 & 54.32$\pm$0.7 & 2.03$\pm$1.4 & & & 17.5$\pm$3.0 & 15.73$\pm$0.8 & 66.17$\pm$8.0 \\

$\textnormal{BCP}$ & & & 65.92$\pm$1.2 & 53.98$\pm$0.8 & 2.15$\pm$0.8 & & & 20.91$\pm$5.9 & 18.64$\pm$0.7 & 62.33$\pm$8.5 \\

$\textnormal{CAML}$ & & & 66.15$\pm$1.4 & 55.01$\pm$0.5 & 2.09$\pm$0.4 & & & 19.68$\pm$6.3 & 17.54$\pm$0.8 & 60.72$\pm$9.3 \\

$\textnormal{DHC}$ & & & 67.19$\pm$0.9 & 55.65$\pm$1.3 & 2.01$\pm$0.7 & & & 25.64$\pm$0.9 & 21.43$\pm$1.1 & 56.82$\pm$4.0 \\

$\textnormal{Cross-ALD}$ & & & 80.61$\pm$1.7 & 67.83$\pm$1.2 & 1.9$\pm$0.4 & & & 26.79$\pm$0.5 & 22.13$\pm$0.8 & 54.25$\pm$2.2 \\

$\textnormal{ACTION++}$ & & & 88.5$\pm$0.9 & 74.32$\pm$1.4 & 0.72$\pm$0.3 & & & 27.41$\pm$0.9 & 24.37$\pm$0.4 & 53.69$\pm$1.2 \\

\rowcolor{red!30} \textbf{DACL (Ours)} & & & \textbf{90.12$\pm$0.4} & \textbf{76.12$\pm$1.6} & \textbf{0.55$\pm$0.2} & & & \textbf{29.88$\pm$0.7} & \textbf{26.54$\pm$1.2} & \textbf{51.32$\pm$1.4} \\ \midrule

$\textnormal{CPS}$ & \multirow{10}{*}{{\textbf{7(10\%)}}} & \multirow{10}{*}{{\textbf{63(90\%)}}} & 85.71$\pm$0.9 & 70.53$\pm$0.6 & 4.00$\pm$0.9 &  \multirow{10}{*}{{\textbf{4(20\%)}}} & \multirow{10}{*}{{\textbf{16(80\%)}}} &  31.46$\pm$2.5 & 28.13$\pm$2.0 & 40.74$\pm$6.9 \\

$\textnormal{MC-Net}$ & & & 86.44$\pm$1.7 & 72.64$\pm$1.5 & 1.84$\pm$0.4 & & & 32.46$\pm$3.3 & 29.64$\pm$1.7 & 43.21$\pm$7.1 \\ 

$\textnormal{URPC}$ & & & 83.37$\pm$0.2 & 69.31$\pm$0.8 & 1.56$\pm$0.2 & & & 25.68$\pm$5.1 & 23.56$\pm$1.2 & 72.74$\pm$15.5 \\

$\textnormal{SS-Net}$ & & & 86.78$\pm$0.7 & 75.12$\pm$1.3 & 1.45$\pm$0.2 & & & 35.08$\pm$0.6 & 32.47$\pm$1.3 & 50.81$\pm$6.5 \\ 

$\textnormal{BCP}$ & & & 87.62$\pm$0.5 & 75.86$\pm$1.2 & 1.46$\pm$0.3 & & & 35.57$\pm$2.9 & 31.65$\pm$1.5 & 36.39$\pm$3.5 \\ 

$\textnormal{CAML}$ & & & 88.48$\pm$0.2 & 76.12$\pm$1.0 & 1.35$\pm$0.3 & & & 37.35$\pm$2.6 & 34.61$\pm$1.2 & 37.42$\pm$2.8 \\

$\textnormal{DHC}$ & & & 87.34$\pm$0.6 & 75.63$\pm$1.0 & 1.37$\pm$0.4 & & & 39.61$\pm$3.9 & 36.45$\pm$1.8 & 34.73$\pm$2.6 \\

$\textnormal{Cross-ALD}$ & & & 87.52$\pm$0.9 & 76.42$\pm$1.4 & 1.6$\pm$0.3 & & & 37.93$\pm$1.1 & 34.76$\pm$1.2 & 37.24$\pm$1.9 \\

$\textnormal{ACTION++}$ & & & 89.4$\pm$1.1 & 78.16$\pm$1.2 & 0.59$\pm$0.4 & & & 39.43$\pm$0.9 & 36.48$\pm$1.2 & 34.19$\pm$1.2 \\
\rowcolor{red!30} \textbf{DACL (Ours)} & & & \textbf{90.91$\pm$0.5} & \textbf{79.12$\pm$1.0} & \textbf{0.38$\pm$0.2} & & & \textbf{41.32$\pm$1.2} & \textbf{37.54$\pm$1.0} & \textbf{32.65$\pm$1.1} \\ \midrule
\end{tabular}}
\caption{Comparison with state-of-the-art methods on the ACDC dataset (with 5\% and 10\% label data) and Synapse dataset (with 10\% and 20\% label data). Metrics reported the mean$\pm$standard results with three random seeds.}\label{table1}
\end{table*}

\noindent\subsubsection{Low-density (sparse) Anchors Sampling.} Density values are used to measure the representativeness of features within a class, indicating the ``hardness" of training for each feature. Features with low-density values indicate less representativeness in the class, \textit{i.e.,} insufficient training for the current model state or being trapped in local optima. Therefore, addressing sparse regions of features will benefit the learning process. For all features of a specific class within a mini-batch, we first estimate their densities based on the corresponding memory bank. They are then ranked by their density values. 
For each class $n$, we select the $N_q$ features with the lowest densities as anchors and store them in $\mathcal{M}^n$:
\begin{equation}\small
\mathcal{M}^n = \left\{ m_i^n : i \in \underset{m_i^n \in \mathcal{V}_n}{\arg \min} \left( d_{m_i^n} \right), \text{ top } N_q \right\},
\end{equation}
where $d_{m_i^n}$ denotes the density value of feature $m_i^n$ and $\arg \min$ selects the indices of the $N_q$ lowest density values in the current mini-batch.

\noindent\subsubsection{High-density (dense) Positive Key Sampling.} We approximate the class center using only high-density features to guide the anchors. During training, a total of $N_p^{+}$ high-density positive keys are sampled from both the mini-batch and the memory bank. While batch features provide fresh and intra-object contrasts, the global memory bank exhibits more comprehensive and diverse category patterns. Therefore, assuming complementarity between these two sets of features can provide robust center estimation. We select the $\frac{1}{2} N_p^{+}$ positive keys with the highest density from the mini-batch and the other $\frac{1}{2} N_p^{+}$ from the memory bank for class $n$ and store them in $\mathcal{P}_{\text{type}}^{n,+}$ as:
\begin{equation}\small
\mathcal{P}^{n,+} = \left\{m_i^n : i \in \underset{\mathbf{p}_{i}^n \in \mathcal{V}_n \cup \mathcal{G}^n}{\arg \max} \left( d_{m_i^n} \right), \text{ top } \frac{1}{2} N_p^{+} \right\},
\end{equation}
where $\theta_{\text{global/local}}^n$ denotes the $N_p^{+}$-th highest intra-class density value in the current batch/memory bank. Next, we calculate the category cluster center as:
\begin{equation}\small
\mathbf{p}_{\text {center}}^{n,+}=\frac{1}{\left|\mathcal{P}^{c,+}\right|} \sum_{\mathbf{p}^{+} \in  \mathcal{P}^{n,+}} \mathbf{p}^{+}.
\end{equation}
\noindent\subsubsection{Negative Key Sampling.} We randomly sample $N_p^{-}$ out-of-class features for each class $n$ from the anchors in the current batch, forming $\mathcal{P}^{n,-}=\left\{\mathbf{p}^{-}\right\}$ to create negative contrast pairs.

\noindent\subsubsection{Positiveness Scores.} We have designed pair-wise positiveness scores to softly measure (in a non-binary form) the relevance between low-density features and positive keys within the same class. For sample pairs ($m^n_i$, $\mathbf{p}_{\text {center}}^{n,+}$), we use the following positiveness score $w_i$ to adjust the contribution of low-density features in contrastive learning:
\begin{equation}\small
w_i=\frac{1}{\gamma_i} \texttt{softmax}\left[l_1\left(m^n_i\right) \times l_2\left(\mathbf{p}_{\text {center }}^{n,+}\right)^{\top}\right], \quad {m}^n_i \in \mathcal{M}^n, \label{coefficients}
\end{equation}
where $l_1(\cdot)$ and $l_2(\cdot)$ are parameter-free identity mapping layers in feature transformation. $\gamma_i$ is a scaling factor to adjust $w_i$.

\noindent\subsubsection{Soft Density-Guided Contrastive Learning.} We utilize a pixel-level contrastive loss to align low-density anchors $m^n_i$ with their the cluster center $\mathbf{p}_{\text {center}}^{n,+}$ approximated by local and global high-density positive keys. This approach is inspired by the SupCon~\cite{khosla2020supervised}. Our methodology includes selective sampling and optimization across all classes within a batch. Soft Density-Guided Contrastive Learning employs the following contrastive loss function, with $\tau$ as the temperature parameter:
\begin{equation}\small
\mathcal{L}_{\text {contra }}=-\sum_{n \in \mathcal{N}} \sum_{m_i \in \mathcal{M}^n} \log \frac{w_i \cdot \exp \left(m_i \cdot \mathbf{p}_{\text {center}}^{n,+} / \tau\right)}{\sum_{\mathbf{p}^{n,-} \in \mathcal{P}^{n,-}} \exp \left(m_i \cdot \mathbf{p}^{n,-} / \tau\right)}. 
\label{13}
\end{equation}

\noindent\textbf{Claim 1.} \textit{Assume we train models using the proposed optimization method, $\mathbf{p}_{\text {center}}^{n,+}$ and $\mathcal{M}^n$ are $n$-th class high-density positive keys and low-density anchor set, respectively. The optimal value of similarity measure $s_i^*$ can be expressed as $\frac{w_{i}}{\sum_{k=1}^{N_q} w_{k}}$, where $w_{i}$ is the corresponding positiveness score for the prototype pair  ($\mathbf{p}_{\text {center}}^{n,+}, \quad {\mathbf{m}}_{i} \in \mathcal{M}^n $) in Eq.~\ref{13}.} 


\section{Experiments and Results}
\noindent\subsection{Dataset and Evaluation Metrics} 
We evaluate our proposed method on four public datasets with different imaging modalities, the Automatic Cardiac Diagnosis Challenge dataset (ACDC)~\cite{bernard2018deep} and Synapse Dataset~\cite{landman20152015}. We evaluate our model and other methods using four criteria on the same testing set for fair comparisons, including the Dice similarity coefficient (Mean Dice), Jaccard (JC), and the average surface distance (ASD).
We stack the results of each testing case in all folds together and report the mean and standard deviation (in parentheses) of each metric. The best-column results are highlighted in red and bold.

\noindent\subsection{Implementation Details}
All models are trained with the SGD optimizer, where the initial learning rate is 0.01, momentum is 0.9 and weight decay is $10^{-4}$. The network converges after 30,000 iterations of training. An exception is made for the first 1,000 iterations, where $\lambda_{\text{cross}}$ and $\lambda_{CL}$ are set to 1 and 0, respectively, which prevents model collapse caused by the initialized prototypes. Empirically, the hyperparameter $N_q$ (the number of anchors per class in each mini-batch) is set to 256. For each anchor, the number of positive keys $N_p^{+}$ and negative keys $N_p^{-}$ are both set to 512. The temperature coefficient $\tau$ in Eq.~\ref{13} is set to 0.4. More details are in the appendix.

\noindent\subsection{Comparisons with State-of-the-Art Methods} 

\begin{figure*}[t]
  \centering
  \begin{tabular}{cc}
    \includegraphics[width=0.95\textwidth]{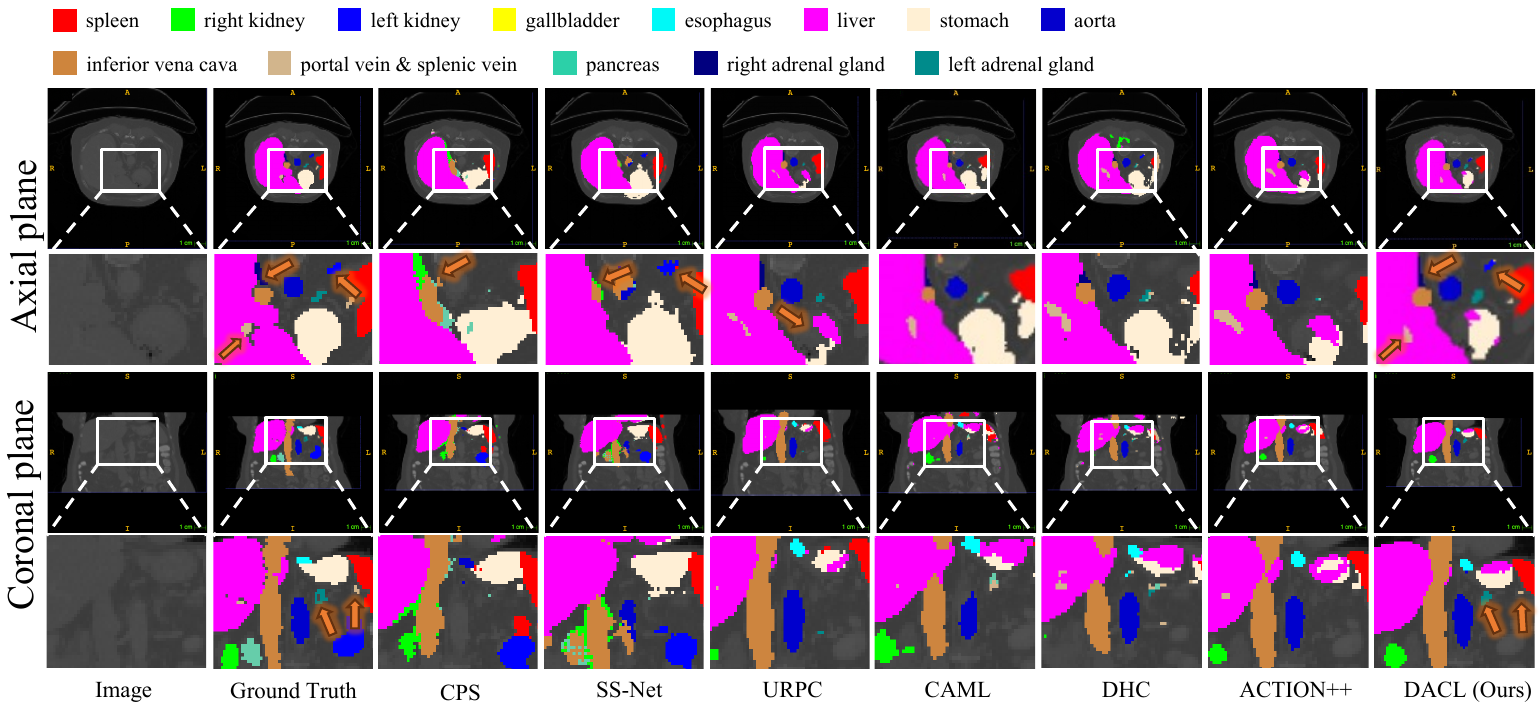} 
  \end{tabular}
  \caption{Visualization of the segmentation results from different methods on the Synapse dataset.}
  \label{visSynapse}
\end{figure*}

We evaluate our proposed model on two datasets using different baseline models~\cite{chen2021semi,wu2021semi,luo2021efficient,wu2022exploring,bai2023bidirectional,gao2023correlation,wang2023dhc,nguyen2023cross,you2023action++} to verify the effectiveness of segmentation. Our method is compared with fully supervised learning (SL) and seven state-of-the-art SSL methods from three categories.

\noindent \subsubsection{ACDC Dataset.} As given in Table~\ref{table1}, DACL surpasses existing methods under all settings and achieves a new state-of-the-art. It also achieves remarkable improvements compared to the second-best results on three partitions (Dice: +1.62\%, 1.51\%; ASD: +0.17mm, +0.21mm). This further demonstrates the effectiveness of our proposed framework, especially in the challenging 5\% labeled data setting where our model significantly outperforms the contrastive learning-based CAML method (Dice: +1.62\%).

\begin{table}[t]
    \centering
    \tiny
    \resizebox{\columnwidth}{!}{
    \begin{tabular}{c|c|c|c|c|c|ccc|ccc}
    \toprule[1pt]
    & PCL  & DA & MS & Bank & $w_i$ & ACDC & Synapse \\
    \midrule I & &  &  & & & 85.71$\pm$0.9 & 18.07$\pm$1.2   \\
    II & \Checkmark &  &  & & & 87.50$\pm$0.8 & 24.32$\pm$0.6   \\
    III & \Checkmark & \Checkmark & & & & 88.60$\pm$0.7 & 25.45$\pm$0.6   \\
    IV & \Checkmark & \Checkmark & \Checkmark & & & 89.30$\pm$0.6 & 26.30$\pm$0.5   \\
    V & \Checkmark & \Checkmark &  \Checkmark &  \Checkmark & & 89.80$\pm$0.5 & 27.12$\pm$0.5   \\
    VI & \Checkmark & \Checkmark &  \Checkmark &  \Checkmark & \Checkmark & \textbf{90.91$\pm$0.5} & \textbf{29.88$\pm$0.7} \\
    \toprule[1pt]
    \end{tabular}}
    \caption{Ablation study on main components of the proposed framework on ACDC and Synapse dataset. PCL: Plain contrastive learning. DA: Density-aware anchor and key sampling at a single scale and in min-batch. MS: Multi-scale density estimation. Bank: Density is estimated with class memory, and positive keys are extended. Positiveness $w_i$: Correlation between low-density features and cluster center.} \label{22ab}
\end{table}

\noindent \subsubsection{Synapse Dataset.} Table~\ref{table1} lists the comparison results on the more challenging Synapse dataset with 14 categories. The data features complex anatomical contrasts, convoluted boundaries, and heterogeneous textures. Under different settings, the improvements of DACL over the state-of-the-art methods are +2.47\% and +1.89\% in Dice demonstrating the outstanding robustness of DACL. The performance improvement of the model is attributed to DACL's effective utilization of geometric structures in the feature space, which enhances intra-class compactness and robustness, particularly in scenarios involving complex anatomical contrasts and heterogeneous textures.

\begin{figure}[t]
  \centering
  \begin{tabular}{cc}
    \includegraphics[width=0.45\textwidth]{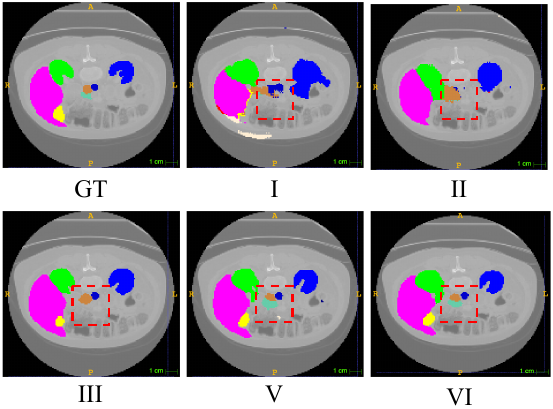} 
  \end{tabular}
  \caption{Qualitative results on Synapse dataset using different components, corresponding to experiments I, II, III, V, and VI in the table~\ref{22ab}.}
  \label{visalb}
\end{figure}

\noindent \subsubsection{Visual Comparison.} As shown in Fig.~\ref{visSynapse}, the baseline method performs poorly in predicting minority classes and blurry boundaries. The segmentation results of DACL consistently give better boundary accuracy compared to other methods, aligning more closely with the ground truths. Similar to ACTION++, DACL can segment tiny objects with complex boundaries (\textit{e.g.,} gallbladder) and large objects with fine structures (\textit{e.g.,} stomach), as indicated by the orange arrows.  This demonstrates DACL's ability to capture semantics and object structures more accurately.

\noindent\subsection{Ablation Study}
We conduct an ablation study on the ACDC and Synapse datasets using 10\% labeled data, evaluating each component's contribution with the Dice score.
\noindent \subsubsection{Density-Aware Contrastive Learning.} In the ablation study, we analyze the contributions of each component, conducting ablation studies on the Synapse dataset. The quantitative results are given in Table~\ref{22ab}, with the CPS~\cite{chen2021semi} model as the baseline. We then introduce plain contrastive learning with a random sampling strategy (Experiment II), which only brings limited gains. In Experiment III, we introduce single-scale density information to guide training (mini-batch density estimation and positive keys sampling), significantly improving the performance by 1.13\% in Dice. In Experiment IV, we introduce multi-scale density-aware nearest neighbor graphs, improving the performance by 0.85\% in Dice. In Experiment V, when estimating density in memory and extending positive samples through memory features, the performance is further improved by 0.82\% in Dice, indicating the importance of leveraging the memory bank in our framework. In Experiment VI, considering the hardness of instances in contrastive learning, we introduce adaptive weights $w_i$ to adjust the low-density features and cluster centers (unlike uniform weight for different features), further improving the performance to achieve the highest Dice score of 29.88$\pm$0.7. 

\noindent \subsubsection{Visualization of Results.} As shown in Fig.~\ref{visalb}, the baseline combines density-guided contrastive learning (III), multi-scale density estimation (IV), and adaptive weights (VI), further improving segmentation accuracy. The model trained with our DACL strategy can eliminate small false positive regions and accurately detect the details of most organs, especially near adherent edges or in low-contrast regions.

\begin{table}[t]
    \centering
    \tiny
    \resizebox{\columnwidth}{!}{
    \begin{tabular}{c|c|c|ccc}
        \toprule[1pt]
        & \hspace{10pt} \textbf{Methods} \hspace{10pt} & \hspace{10pt} \textbf{ACDC} \hspace{10pt} & \hspace{10pt} \textbf{Synapse} \hspace{10pt} \\
        \midrule
        & Baseline & 85.71$\pm$0.9 & 18.07$\pm$1.2 \\ \midrule
        \multirow{4}{*}{{\textbf{Class-wise}}} & + ED     & 87.50$\pm$0.8 & 22.88$\pm$0.7 \\
        & + IFVD    & 88.60$\pm$0.7 & 25.00$\pm$0.6 \\
        & + PLCL   & 89.30$\pm$0.6 & 26.50$\pm$0.5 \\ 
        & + Ours    & 90.50$\pm$0.5 & 27.50$\pm$0.5 \\\midrule
        \multirow{4}{*}{{\textbf{Pixel-wise}}} & + InfoNCE & 87.00$\pm$0.8 & 23.00$\pm$0.7 \\
        & + PLCT  & 88.20$\pm$0.7 & 25.50$\pm$0.6 \\
        & + SCM  &  89.10$\pm$0.6 & 26.70$\pm$0.5 \\
        & + Ours   & \textbf{90.91$\pm$0.5} & \textbf{29.88$\pm$0.7} \\
        \toprule[1pt]
    \end{tabular}}
    
    \caption{Comparison of different Class-wise and Pixel-wise losses on the ACDC and Synapse dataset.}\label{loss}
    
\end{table}

\noindent \subsubsection{Comparison of Contrastive Losses.} 
To verify the capability of density constraints for class reasoning, we replace them with several standard class-wise and pixel-wise losses, including ED~\cite{li2022category}, IFVD~\cite{wang2020intra}, PLCL~\cite{alonso2021semi}, SCM~\cite{miao2023sc}, and PLCT~\cite{chaitanya2023local}. As given in Table~\ref{loss}, our proposed pixel-wise contrastive learning module significantly outperforms other these methods, demonstrating that DACL enhances semantic discriminative ability in the feature space, whereas other methods lack semantic distinction.

\noindent \subsubsection{Effectiveness of positiveness metrics.}

\begin{figure}[t]
  \centering
  \begin{tabular}{cc}
    \includegraphics[width=0.45\textwidth]{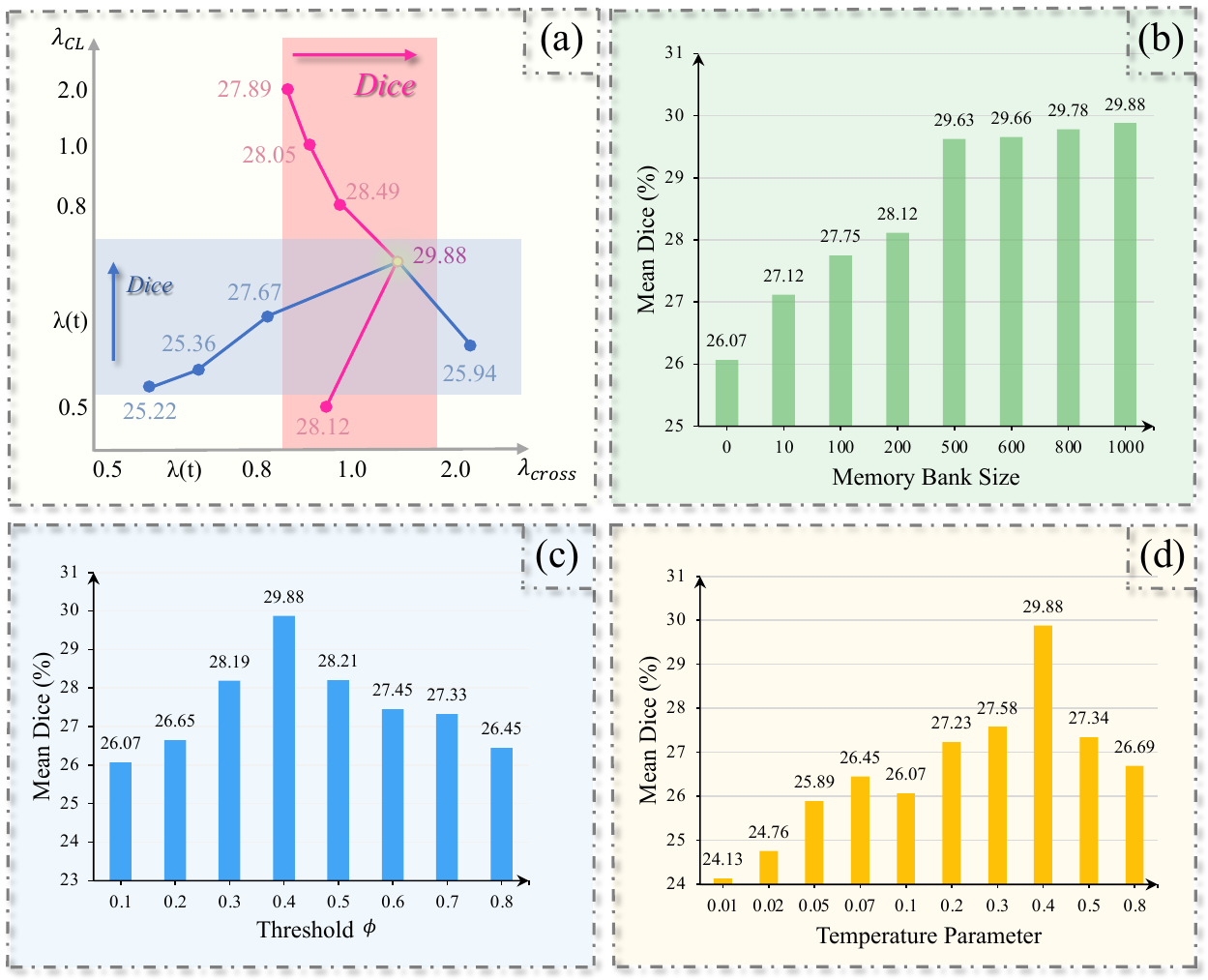} 
  \end{tabular}
  \caption{Mean Dice performances on the Synapse dataset with different hyperparameters. Performances with various (a) $\lambda_{cross}$ and $\lambda_{CL}$ in Eq.~\ref{eq1}, (b) Memory bank size, (c) threshold $\phi$ for generating the 0-1 seed mask in Eq.~\ref{tau}, and (c) temperature hyperparameters $\tau$ in Eq.~\ref{13}.}
  \label{Hyperparameters}
\end{figure}

We evaluate the effects of $\lambda_{cross}$ and $\lambda_{CL}$, as well as the Gaussian warming-up function $\lambda(t)$. As shown in Fig.~\ref{Hyperparameters} (a), we first evaluate the effect of $\lambda_{CL}$ when $\lambda_{cross} = 1.0$ (in pink). When the value of $\lambda_{CL}$ is low (\textit{i.e.,} $\lambda_{CL} = 0.5$), our density-guided contrastive learning positively impacts optimization. However, we find higher values to be detrimental, possibly because the overall loss is overwhelmed by the high uncertainty of the unlabeled data, hindering the training process. The best performance is achieved when choosing $\lambda(t)$ as $\lambda_{CL}$, as it can adaptively increase the weight according to training iterations. Next, we fixed $\lambda_{CL} = \lambda(t)$ and used different values of $\lambda_{cross}$ (in blue). Overall, $\lambda_{cross} = 1$ achieves the best results. As shown in Fig.~\ref{Hyperparameters} (b), compared to not using the Memory bank, Mean Dice increases by at least 1.05\%. Also, we observe a positive correlation between Memory bank size and performance, especially at length 1000. As shown in 
Fig.~\ref{Hyperparameters} (c), demonstrates stable performance across different 0-1 seed mask thresholds $\phi$, indicating insensitivity to this hyperparameter.
Fig.~\ref{Hyperparameters} (d) indicates an optimal temperature $\tau$ of 0.4, consistently outperforming the baseline.

\section{Conclusion}
In this paper, we propose a novel framework named DACL for semi-supervised multi-organ segmentation. Our key idea is to mine effective supervision from the geometry of clusters. To this end, two novel modules: Feature Density-Aware module (FDA) and Soft Density-guided Contrastive Learning (SDCL) are proposed to mine effective neighborhood information by tackling sparse regions inside the feature space. Extensive experimental results on two public multi-organ segmentation datasets demonstrate that we outperform previous state-of-the-art results by a large margin without extra computational load required.

\section{Acknowledgments}
This research was supported by an Australian Government Research Training Program (RTP) Scholarship awarded to Zhongxing Xu.

\bibliography{aaai25}

\end{document}